\documentclass[final,5p,times,twocolumn]{elsarticle}

\usepackage{amssymb}
\usepackage{amsmath,bm}
\usepackage{algorithm}
\usepackage{algpseudocode}
\usepackage{caption}
\usepackage{graphicx} 
\usepackage{url}

\journal{TBD}

\date{April 2024}

\usepackage{amsfonts}
\usepackage{amsthm}
\usepackage{amsfonts}
\usepackage{amsmath}
\usepackage{amscd}
\usepackage{bbm}
\usepackage{booktabs}
\usepackage[utf8]{inputenc}
\usepackage[T1]{fontenc}
\usepackage[mathscr]{eucal}
\usepackage{indentfirst}
\usepackage{graphicx}
\usepackage{xcolor}
\usepackage{hyperref}
\usepackage{graphics}
\numberwithin{equation}{section}
\usepackage{hyperref}
\usepackage{algorithm}
\usepackage{algpseudocode}
\usepackage{color, colortbl}

\usepackage[numbers]{natbib}
\usepackage[flushleft]{threeparttable}
\usepackage{multirow}
\usepackage{subcaption}

\definecolor{LightCyan}{rgb}{0.88,1,0.92}

\algrenewcommand\algorithmicrequire{\textbf{Input:}}
\algrenewcommand\algorithmicensure{\textbf{Output:}}

\begin{document}

\begin{frontmatter}
 
\title{DynBERG: Dynamic BERT-based Graph neural network for  financial fraud detection}

\author[inst1,inst3]{Omkar Kulkarni}
\ead{f20201172@goa.bits-pilani.ac.in}  

\author[inst2,inst3]{Rohitash Chandra}
\ead{rohitash.chandra@unsw.edu.au} 
\affiliation[inst2]{Transitional Artificial Intelligence Research Group, School of Mathematics and Statistics, UNSW Sydney, Australia}

\affiliation[inst3]{Centre for  Artificial Intelligence and Innovation, Pingala Institute, Sydney, Australia}

\affiliation[inst1]{organization={Department of Economics and Finance},
            addressline={BITS Pilani K.K. Birla Goa Campus}, 
            state={Goa},
            postcode={403726}, 
            country={India}}

\begin{abstract}
Financial fraud detection is critical for maintaining the integrity of financial systems, particularly in decentralised environments such as cryptocurrency networks. Although  Graph Convolutional Networks (GCNs) are widely used for financial fraud detection, graph Transformer models such as Graph-BERT are gaining prominence due to their Transformer-based architecture, which mitigates issues such as over-smoothing. Graph-BERT is designed for static graphs and primarily evaluated on citation networks with undirected edges. However, financial transaction networks are inherently dynamic, with evolving structures and directed edges representing the flow of money. To address these challenges, we introduce DynBERG, a novel architecture that integrates Graph-BERT with a Gated Recurrent Unit (GRU) layer to capture temporal evolution over multiple time steps. Additionally, we modify the underlying algorithm to support directed edges, making DynBERG well-suited for dynamic financial transaction analysis. We evaluate our model on the Elliptic dataset, which includes Bitcoin transactions, including all transactions during a major cryptocurrency market event, the Dark Market Shutdown. By assessing DynBERG's resilience before and after this event, we analyse its ability to adapt to significant market shifts that impact transaction behaviours. Our model is benchmarked against state-of-the-art dynamic graph classification approaches, such as EvolveGCN and GCN, demonstrating superior performance, outperforming EvolveGCN before the market shutdown and surpassing GCN after the event. Additionally, an ablation study highlights the critical role of incorporating a time-series deep learning component, showcasing the effectiveness of GRU in modeling the temporal dynamics of financial transactions. We provide an open-source code implementation for applying DynBERG to other dynamic graph anomaly detection problems in financial and other domains.
\end{abstract}



\end{frontmatter}

\section{Introduction}

Real-world datasets can be expressed in terms of graphs, which capture both the node attributes and their complex relationships represented by edges. Examples of applications of graphs to real-world datasets include brain imaging data \citep{Yu2018GraphTheoryBrain}, social media\citep{ugander2011anatomyfacebooksocialgraph} and bio-medical molecules\citep{Huber2007GraphsMolecularBiology}. Traditional machine learning algorithms take feature vectors as input and, therefore, cannot be used directly with graph data. This creates the need to preprocess the node and edge features in a graph, making the learning of graph representations an important task \citep{laclau2024surveyfairnessmachinelearning}.

Several deep learning models have been developed in the past, with a focus on static graphs that remain unchanged over time \citep{GraRep, kipfwellingGCN,graphattentionnetworks, graphunets}. However, in real-life applications, graphs often evolve, making them dynamic in nature, with nodes and edges changing over time. For example, relationships in social networks evolve, so the graphs representing a user's connections should be updated as their social relationships change over time. Likewise, citation networks continuously expand as new research papers are published, referencing earlier works. As a result, the influence of an article and sometimes even its classification changes over time. Therefore, it is essential to update node embeddings to capture these evolving relationships. In the financial domain, users engage in transactions continuously, causing the nature of their accounts to change based on their interactions with other users. Early detection of illicit transactors is crucial for maintaining the integrity of financial networks, as attempted by deep learning models \citep{dlfrauddetect, dlfrauddetect2}, especially in the cryptocurrency space, where decentralised cryptocurrencies such as  Bitcoin allow users to remain pseudo-anonymous \citep{bitcoin}. Therefore, studying such temporally evolving financial networks is essential for identifying illicit transactors.

Prior work related to graph anomaly detection in the financial domain has focused mostly on static financial graphs \citep{graphanomalystatic, graphanomalystatic2, graphanomalystatic3, graphanomalystatic4} with little focus on dynamic evolving graphs \citep{graphanomalydynamic, graphanomalydynamic2}. The underlying framework for most of these models is that of Graph Convolutional Networks (GCN)\citep{kipfwellingGCN} and Evolving Graph Convolutional Networks (EvolveGCN)\citep{evolvegcn}. Both of these frameworks use the concept of convolution or message-passing in a graph context. Message passing has associated problems such as over-smoothing \citep{oversmoothing}, which causes inaccurate results.  Li et al. \citep{oversmoothing} have shown that as the model architecture goes deeper for GNNs (Graph Neural Networks) based on the concept of graph convolutional operators \citep{gco}, and reaches a certain limit. The node representations learned from such deep models tend to be over-smoothed and also become indistinguishable. Hence, to overcome this problem, Transformer-based graph neural network model called `Graph-BERT'\citep{graphbert} was introduced to address two key challenges. First, by utilising a Language Models  such as  BERT (Bidirectional Encoder Representation from Transformers) \citep{bert1} to extract context embeddings. This works on the underlying concept of Transformers instead of convolutional layers, which effectively mitigates the issue of over-smoothing. Second, it introduced a method to partition a large input graph into smaller, fixed-size subgraphs, significantly reducing the computational resources needed to run the model. 

However, a key limitation of the Graph-BERT \citep{graphbert} model is that it was designed for static graphs and primarily tested on citation network datasets such as Cora, Citeseer, and Pubmed. Additionally, it was developed for graphs with undirected edges, making it less suitable for financial transaction graphs, which are typically dynamic and feature directed edges representing the flow of money. The novelty of our work lies in introducing DynBERG, an architecture that integrates Graph-BERT with a Gated Recurrent Unit (GRU) \citep{gru} layer to capture temporal evolution across subgraphs over multiple timestamps. We also modify the underlying algorithm to support directed edges, making our model well-suited for financial transaction networks that evolve over time. We evaluate DynBERG on the Elliptic dataset \citep{elliptic}, which contains Bitcoin transactions, including a major cryptocurrency market event called `Dark Market Shutdown'. Our model's resilience is tested before and after this event to assess its performance in response to significant market shifts that impact transaction behaviours in the Bitcoin network \citep{elliptic}. Furthermore, we compare DynBERG against state-of-the-art dynamic graph classification models, such as EvolveGCN and GCN. 

The rest of the paper is organised as follows. Section 2 provides a background on related methods, and Section 3 presents the data pre-processing and the proposed method. 
Section 4 presents experiments and the results, Section 5 discusses the results, and Section 6 concludes with future research directions.

\section{Related work}

\subsection{Graph neural networks}

Dynamic graph methods are extensions of static graph approaches with a greater focus on incorporating mechanisms to handle temporal dynamics and update schemes. For example, in matrix-factorisation-based approaches, the eigenvectors of the graph Laplacian matrix are used as the node embeddings \citep{matrixfact1}. Hence,  Li et al. \citep{matrixfact2} use eigenvectors from prior time steps to update the new ones instead of freshly computing them. The advantage of such methods is their computational efficiency. Further, random walk-based approaches focus on maximising the probabilities of sampled random walks \citep{randomwalk2}. Nguyen et al.\citep{randomwalk1} build on this concept by enforcing that the walks follow the temporal order.

Moving on to deep learning-based approaches, we get an abundance of supervised and unsupervised approaches. Goyal
et al. \citep{deeplearning1} proposed a model DynGEM, that is an autoencoding approach that focussed on minimisation of the reconstruction loss along with the distance between connected nodes in the embedding space. An important feature of DynGEM was that the autoencoder learned from the past time step to initialise the training in the following time step. A widely used approach for modeling dynamic graphs involves point processes, which operate in continuous time. Trivedi et al.\citep{deeplearning2} and Trivedi et al. \citep{deeplearning3} introduced two models Know-Evolve and DyRep, respectively, that model the occurrence of an edge as a point process and use a neural network with the parameterized intensity function. Zuo et al.\citep{deeplearning4}'s model HTNE uses the Hawkes process with an additional attention mechanism to determine the influence of historical neighbours with current neighbours of a node. Due to their continuous nature, these methods are particularly effective for event time prediction.

The more relevant set of approaches considers GNNs in combination with recurrent network architectures such as  LSTM/GRU. The former digests the graph information while the latter handles the dynamism. GCN's\citep{kipfwellingGCN} are the most explored GNNs in this context. Seo et al. \citep{deeplearning5} introduced a model, GCRN, to obtain node embeddings, which were then fed to an LSTM that studied the dynamism. This idea was also adopted in WD-GCN/CD-GCN introduced in Manessi et al. \citep{deeplearning6} that modified graph convolutional layers by adding a skip connection. Furthermore, the model EvolveGCN proposed by Pareja et al. \citep{evolvegcn} uses GCN combined with RNN, but the GCN parameters are not trained anymore; only RNN parameters are trained that compute the GCN parameters. In this way, the model size does not increase with the number of time steps, and the model remains as manageable as an RNN. 

However as pointed out by Motie and Raahemi \citep{reviewpaperfin1}, all these models use the underlying convolutional mechanism, i.e. message passing mechanism which suffers from oversmoothing, and there is a need to explore the application of transformers with graph neural networks for real-world applications in financial fraud detection. Zhang et al. \citep{graphbert} introduced a model Graph-BERT, for node classification in static graphs with undirected edges, wherein the problem of oversmoothing was solved due to the application of the transformer-based LLM architecture BERT that was used to extract contextual embeddings. The authors also introduced fixed-size subgraph batching wherein `top-k' similar nodes to a particular node were included in its subgraph, where the degree of similarity was determined by the PageRank algorithm. As explained in the subsequent sections, this subgraph batching allowed large input-sized graphs to be broken down into smaller subgraphs, improving the computational efficiency of the model. However, the application of this model or extension of this model's architecture to dynamic graphs with directed edges has not yet been explored.

We did not find any such study that applies Graph Transformer models, such as Graph-BERT, and integrates them with other recurrent neural network architectures to extract time information from dynamic graphs, with a particular application to node classification in dynamically evolving financial graphs for fraud detection.

\section{Methodology}


\subsection{Data}
\label{sec:data}

\begin{figure*}[htbp!]
  \centering
  \includegraphics[width=0.95\linewidth]{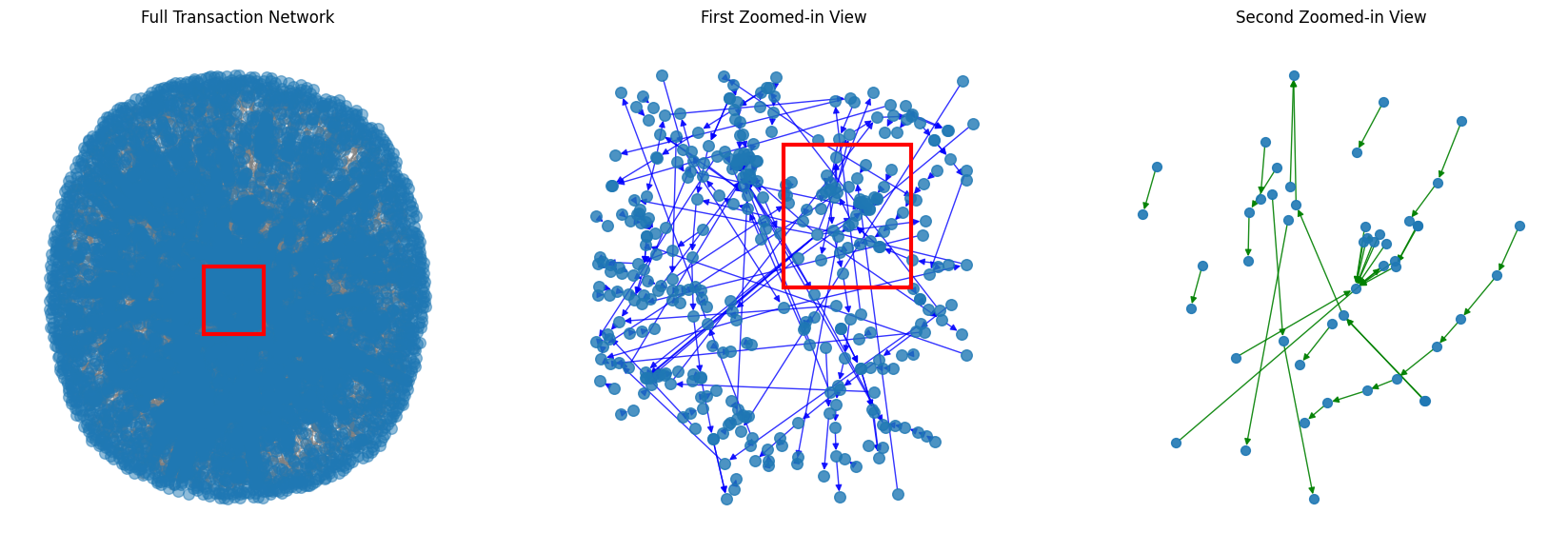}
  \captionof{figure}{Transaction network for Day 1}
  \label{fig:graph1}
\end{figure*}

\begin{figure}[htbp!]
  \centering
  \includegraphics[width=0.95\linewidth]{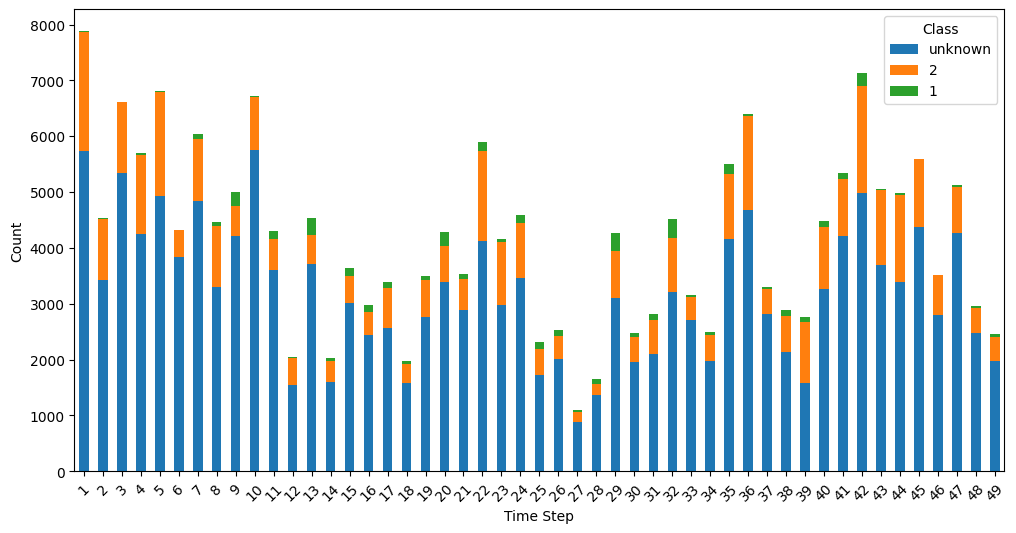}
  \captionof{figure}{Number of nodes vs Time step}
  \label{fig:data1}
\end{figure}

\begin{figure}[htbp!]
  \centering
  \includegraphics[width=0.95\linewidth]{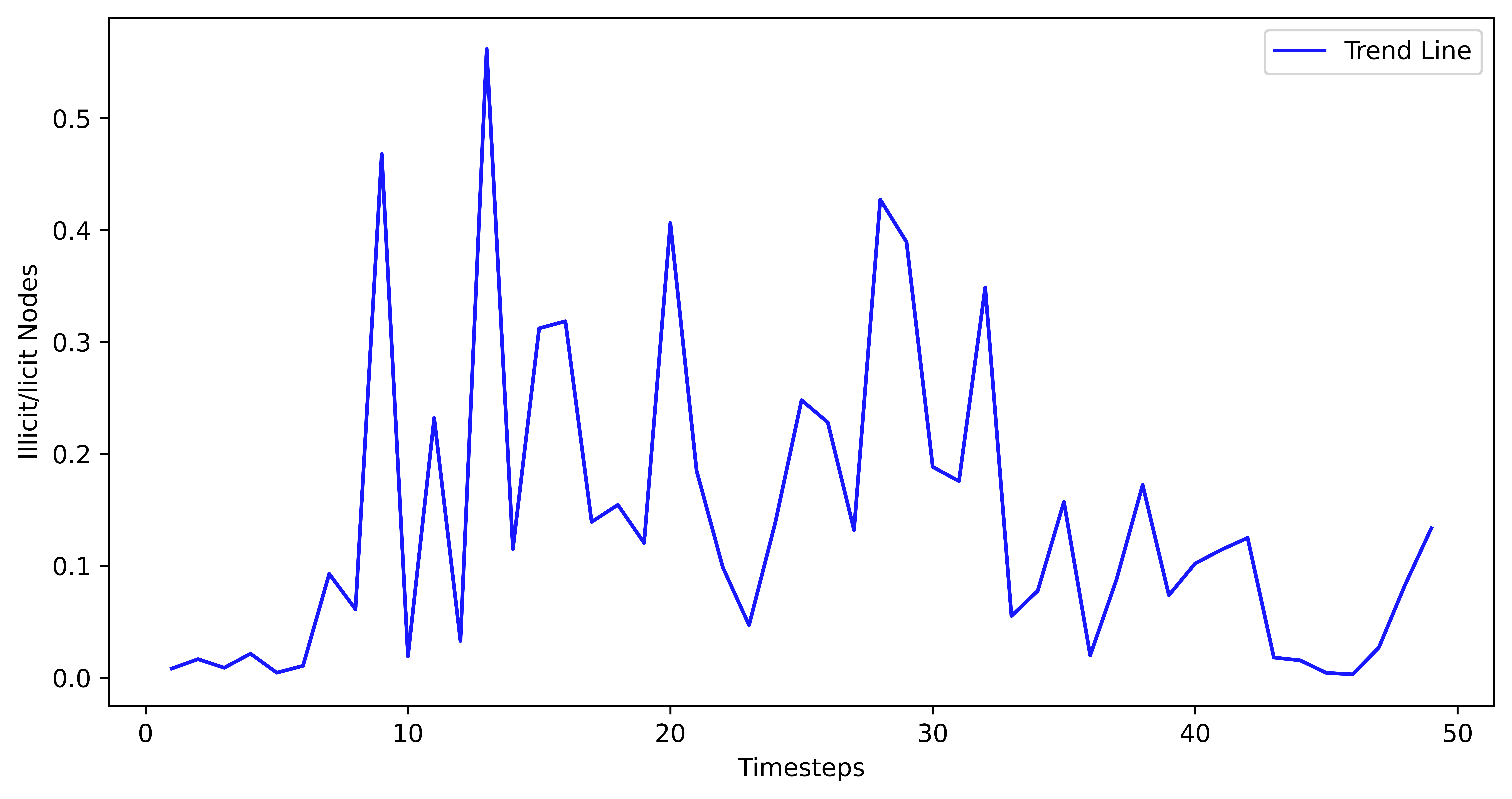}
  \captionof{figure}{Fraction of illicit vs licit nodes at different time steps in the dataset}
  \label{fig:data2}
\end{figure}

We use a publicly available Bitcoin transaction dataset called `Elliptic'. This dataset contains bitcoin transactions for 49 time steps, where each time step comprises a single connected component of transactions that appeared on the blockchain within less than three hours of each other. There are no edges connecting different time steps, and every time step is separated by a duration of two weeks. The 43rd time step in the dataset is recorded just after the dark market shutdown. There are a total of 203,769 nodes (i.e. transactors) and 234,355 edges(i.e. transactions) in this dynamic graph dataset. Approximately 20\% of the nodes have been mapped to real entities and have been classified as licit (exchanges, wallet providers, miners, licit services, etc.) versus illicit (scams, malware, terrorist organisations, ransomware, Ponzi schemes, etc.) categories. Every node is associated with 166 features, where the first 94 represent local information about the transaction, such as time step, transaction fee, output volume, and the remaining 72 represent aggregated features, which are obtained by aggregating transaction information related to the neighbouring nodes. Figure \ref{fig:data1} shows the number of nodes per class vs time steps, while Figure \ref{fig:data2} shows the fraction of illicit vs licit nodes per time step. Figure \ref{fig:graph1} shows the transaction network corresponding to day 1. The subplot on the left is the overall transaction network, while the two subplots on the right expand upon the part of the plot inside the rectangular boxes to show a zoomed-in version of the transactions.

\begin{figure*}[htbp!]
  \centering
  \begin{subfigure}[t]{0.48\linewidth}
    \centering
    \includegraphics[width=\linewidth]{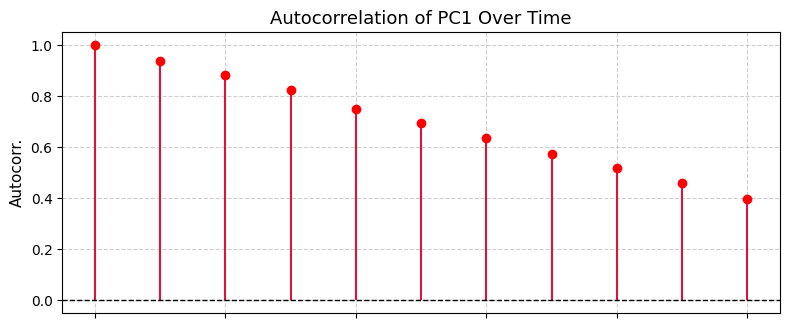}
    \caption{Principal Component 1}
    \label{fig:acfplot1}
  \end{subfigure}
  \hfill
  \begin{subfigure}[t]{0.48\linewidth}
    \centering
    \includegraphics[width=\linewidth]{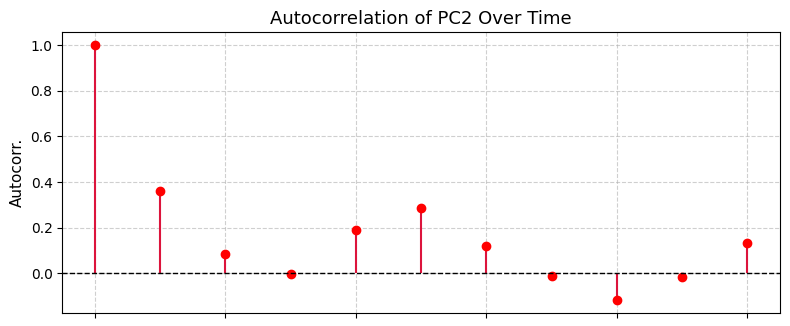}
    \caption{Principal Component 2}
    \label{fig:acfplot2}
  \end{subfigure}
  \caption{ACF plots for the first two principal components.}
  \label{fig:acfplots}
\end{figure*}

\begin{figure*}[htbp!]
  \centering
  \includegraphics[width=0.75\linewidth]{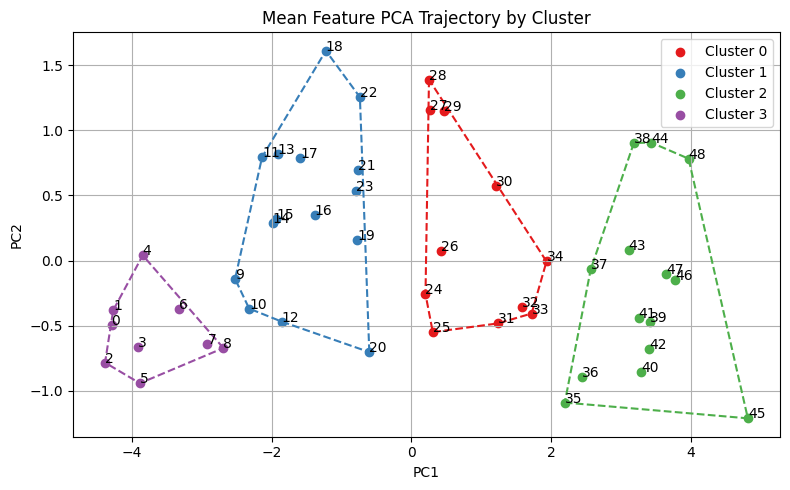}
  \captionof{figure}{Mean PCA cluster plot}
  \label{fig:pcacluster}
\end{figure*}

We further perform a PCA-based clustering and an autocorrelation function analysis to investigate the temporal dynamics of the data, assessing the extent to which a deep learning model can effectively capture this temporal structure. We utilise the top-2 principal components of the 166 features in our data across different time steps. The clusters of the PCA components across different time steps are shown in Figure \ref{fig:pcacluster}. PCA values corresponding to adjacent time steps tend to belong to the same cluster, indicating that temporal information is encoded within the node features. Furthermore, we present the autocorrelation functions for both principal components in Figure
 \ref{fig:acfplots}. The consistently high autocorrelation values across all considered time lags for both components further support the presence of temporal information encoded within the node features.



\subsection{DynBERG}
\label{sec:dynberg}

The graph transformer part of our model DynBERG is motivated by Graph-BERT's subgraph batching and graph transformer model architecture. \textcolor{black}{The framework addresses node classification on dynamic graphs by combining structural and temporal modelling. At each timestep, the graph is partitioned into subgraphs to facilitate efficient batching and capture localised structures. A transformer-based encoder processes these subgraphs, incorporating edge directionality to produce context-sensitive representations. To strengthen these embeddings, a node reconstruction pre-training step is performed before fine-tuning. The resulting subgraph representations are then integrated with a GRU layer, which models temporal dependencies across graph snapshots and yields dynamic node embeddings suitable for classification.} 

\subsubsection{Subgraph batching}
\label{sec:subgraph}

In order to break down a complex, large input graph into multiple fixed-sized smaller graphs, we use subgraph batching introduced by Zhang et al. \citep{graphbert}. In this approach, we compute a graph intimacy matrix represented by $S \in \mathbb{R}^{|\nu| \times |\nu|}$; where $\nu$ represents the set of nodes in the input graph and $|\nu|$ represents the graph size (number of nodes) \textcolor{black}{as shown in figure \ref{fig:framework} where $|\nu|$ = 12}. In the graph intimacy matrix, $S(i,j)$ represents the intimacy score between node $\nu_i$ and $\nu_j$. We compute  matrix $S$ using the pagerank algorithm as defined by $S = \alpha \cdot [I - (1 - \alpha) \cdot \bar{A}]^{-1}$, where  $\alpha \in [0,1]$. $\bar{A}$ represents the row normalised adjacency matrix, where $\bar{A} = D^{-1} \cdot A$, $A$ represents the adjacency matrix of the input graph and $D$ represents the diagonal matrix where $D(i,i) = \sum_j A(i, j)$. The difference between the adjacency matrix in this paper and  \textcolor{black}{Zhang} et. al \citep{graphbert} is in their implementation as their adjacency matrix have been symmetrised, thus they have assumed that the input graph contains undirected edges. Therefore, they use the normalisation $\bar{A} = D^{-1/2} \cdot A \cdot D^{1/2}$ while the normalisation used in this paper, i.e. $\bar{A} = D^{-1} \cdot A$ is motivated by the directed graph neural network proposed by \textcolor{black}{Shi} et al. \citep{normadjpaper} where the edges are directed, hence the adjacency matrix is not symmetric.

After computing the graph intimacy matrix, we select the top-k most intimate nodes to a target node $\nu_i$ and include them in its subgraph. Hence, every node in the large input graph has an associated subgraph comprising $k$ other intimate nodes; thus, every subgraph has a size of $k$+1 nodes. These subgraphs, calculated from a large input graph, are used in the further steps, hence reducing the computational power required to run deep learning models that otherwise would have to process the large graphs.

\subsubsection{Graph Transformer-based Encoder}
\label{sec:graphtrans}

For every node $\nu_j \in V_i$, where $V_i$ denotes the set of nodes in the subgraph for a node $\nu_i$, the embedding of the raw feature vector $x_j$ (with dimension $d_x \times 1$) can be denoted as

\begin{equation}
    e_j^{(x)} = \text{Embed}(x_j) \in \mathbb{R}^{d_h \times 1}
    \label{eq:embedding}
\end{equation}

$Embed(\dot)$ function in Equation \eqref{eq:embedding} summarises the raw feature vector into an embedding vector of dimension ($d_h \times 1$).

Thus, the input to the transformer can be represented as $h_j^{(0)} = e_j^{(x)}$ corresponding to node $\nu_j$ in a subgraph for node $\nu_i$. The input vectors for all nodes in a subgraph for node $\nu_i$ can be represented in the form of a matrix $H^{(0)} = [h_i^{(0)},h_{i,1}^{(0)},h_{i,2}^{(0)},h_{i,3}^{(0)},h_{i,4}^{(0)}......,h_{i,k}^{(0)}]^T \in \mathbb{R}^{(k+1) \times {d_h}}$. The Graph-Transformer-based encoder, as introduced \textcolor{black}{in equation \eqref{eq:graphtransformer}} updates the nodes' representations iteratively with multiple layers ($D$ layers), and the output by the l$th$ layer can be denoted as
\begin{equation}
H^{(l)} = \text{G-Transformer} \left( H^{(l-1)} \right)  \\
        = \text{softmax} \left( \frac{QK^\top}{\sqrt{d_h}} \right) V + \text{G-Res} \left( H^{(l-1)}, X_i \right)
\label{eq:graphtransformer}
\end{equation}

\noindent where

\begin{equation}
    \begin{cases}
    Q = H^{(l-1)} W_Q^{(l)} \\
    K = H^{(l-1)} W_K^{(l)} \\
    V = H^{(l-1)} W_V^{(l)}.
    \end{cases}
\label{eq:graphtransformerDefinition}
\end{equation}

In Equation \eqref{eq:graphtransformerDefinition}, $W_Q^{(l)},W_K^{(l)},W_V^{(l)} \in \mathbb{R}^{d_h \times d_h}$ denote the trainable weights. $\text{G-Res}(H^{(l-1)},X_i)$ in Equation \eqref{eq:graphtransformer} denotes the graph residual term introduced by Zhang and Meng \citep{gres} and $X_i$ denotes the raw features of all nodes in the subgraph corresponding to node $\nu_i$, $X_i = [x_i,x_{i,1},x_{i,2},x_{i,3},x_{i,4}......,x_{i,k}]^T \in \mathbb{R}^{(k+1) \times d_x}$. Thus, the learning process of the graph-Transformer function described above can be denoted as

\begin{equation}
    \begin{aligned}
        H^{(0)} &= \begin{bmatrix} h^{(0)}_i, & h^{(0)}_{i, 1}, & \cdots & h^{(0)}_{i, k} \end{bmatrix}^T \\
        H^{(l)} &= \text{G-Transformer} \left( H^{(l-1)} \right), \quad \forall l \in \{1, 2, \dots, D\} \\
        z_i &= \text{Fusion} \left( H^{(D)} \right).
    \end{aligned}
\label{eq:maingraphtrans}
\end{equation}

The Transformer model defined in Equation \eqref{eq:maingraphtrans} is different from the conventional Transformer used for NLP applications, which focuses on learning the representations of all the input tokens. In this paper, only the representation of the target node in a subgraph or the node to which the subgraph corresponds has to be learned. The Fusion ($\cdot$) in the \textcolor{black}{Equation \eqref{eq:maingraphtrans}} will give an average of all the nodes in the input list, which defines the final state of the target node $\nu_i$, i.e. $z_i \in \mathbb{R}^{d_h \times 1}$. Both vector $z_i$ and $H^{(D)}$ are further input to a fully connected layer that is designed according to the task to be conducted, i.e. pre-training or node classification. Different tasks will have different learning objectives (loss functions).

\subsubsection{Pre-training: Node Raw Attribute Reconstruction}
\label{sec:pretrain}

The goal of pretraining our graph-transformer is to train it to learn the raw attributes of the target node in a subgraph. After pre-training, the transformer can be used for further downstream tasks, such as node classification, which involves fine-tuning of hyperparameters. We have the learned representation $z_i$ corresponding to the subgraph for the target node $\nu_i$. We reconstruct the raw attributes for the target node $\nu_i$ based on $z_i$ using a fully connected layer (denoted by FC($\cdot$), we use an activation function if required), $\hat{x}_i = \text{FC}(z_i)$. To ensure that the learned representation $z_i$ can capture the raw attribute information of the node $\hat{x}_i$ compared to the raw characteristics of the node $x_i$, we define the loss term of reconstruction of the raw attribute of the node as

\[
l_1 = \frac{1}{|\nu|}\sum_{v_i \in V} \left\| x_i - \hat{x}_i \right\|_2
\]

\subsubsection{Dynamic BERG: Fine-tuning for node classification}
\label{sec:finetuning}

 We use a GRU layer to introduce dynamism to the BERG unit. We update the hidden states of the GRU layer using the average-pooled vector of all the final states of the Graph Transformer-based encoder described in Section \ref{sec:graphtrans} corresponding to subgraphs within a particular timestep. Thus, $Z_p = \{z_{1,p}, z_{2,p},....z_{n,p}\}$, where $Z_p$ denotes the set of final states of the encoder model corresponding to $n$ number of subgraphs in the $p$th time step, where $n$ or the total no. of subgraphs in a timestep is variable since there are variable number of nodes for every timestep. The hidden state of the GRU layer corresponding to the $p$th timestep denoted by $HS_p$ is updated as $HS_p$=GRUcell$(AveragePooling(Z_p), HS_{p-1})$, where $AveragePooling(\cdot)$ calculates the mean of all vectors in the given set and $HS_{p-1}$ denotes the hidden state from the $(p-1)$th timestep, thus $HS_p$ is of dimensions $d_h \times 1$.

\begin{figure*}[htbp!]
  \centering
  \includegraphics[width=0.7\linewidth]{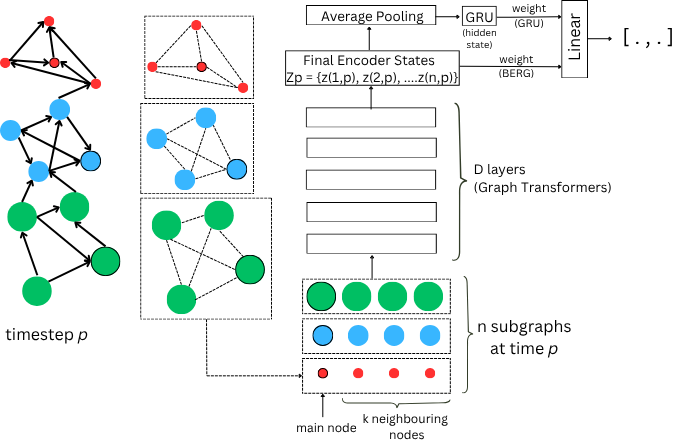}
  \captionof{figure}{Framework: Figure shows the architecture of the model, taking an example of a graph at timestep `p'. Every graph at timestep `p' is divided into `n' subgraphs using the PageRank algorithm, where every subgraph has k+1 nodes. These are fed into the `D' layers of graph transformers, which output the final encoder states. These final encoder states are average-pooled and used to update the hidden state of the GRU layer, which keeps track of the time-related features of the dynamic financial graph. The hidden state of the GRU layer and the final encoder states from the graph transformer layers are used to make class-level predictions.}
  \label{fig:framework}
\end{figure*}

Finally, the node classification for every subgraph in a timestep is performed using a weighted average of the encoder final state corresponding to the particular subgraph in a timestep and the GRU hidden state corresponding to the particular timestep. Hence output of the node classification is $\hat{y}_i = softmax(FC(w_{BERG}z_{i,p} + w_{GRU}HS_p))$ where $z_{i,p}$ corresponds to the encoder final state corresponding to the particular subgraph for target node $\nu_i$ in a timestep $p$ \textcolor{black}{as shown in figure \ref{fig:framework}}. Further, we use the cross-entropy loss function between the predicted labels and the true labels.

\section{Results}


\subsection{Hyperparameter Tuning: Subgraph batch size}
\label{sec:hyperparamtuning}


The subgraph batch size parameter represents the number of nodes to be included in a subgraph which intuitively represents the size of the network closely related to a particular node to classify it as a licit or illicit node. We optimise this parameter and choose the size that gives the best illicit F1 and Micro-F1 scores. According to Table \ref{tab:tuningsubgraphsize}, a subgraph batch size of 11 gives the best illicit F1 and micro F1 hence, we use this value in our further results analysis. 
\begin{table}[htbp]
    \centering
    \begin{tabular}{ccc} 
        \toprule
         k & Illicit F1 & Micro-F1 \\
        \midrule
        3 & 0.5741 & 0.9505 \\
        4 & 0.5766 & 0.9491 \\
        5 & 0.5817 & 0.9488 \\
        6 & 0.5823 & 0.9505 \\
        7 & 0.5967 & 0.9513 \\
        8 & 0.5832 & 0.9504 \\
        9 & 0.5690 & 0.9492 \\
        10 & 0.5977 & 0.9531 \\
        11 & 0.6126 & 0.9606 \\
        12 & 0.5987 & 0.9533 \\
        13 & 0.5997 & 0.9521 \\
        14 & 0.5873 & 0.9513 \\
        15 & 0.5738 & 0.9473 \\
        \bottomrule
    \end{tabular}
    \caption{Illicit and Micro F1 scores for different values of the number of neighbouring nodes of a target node in a subgraph. The highest illicit F1 score is for a subgraph with 11 neighbouring nodes, hence a value of $k=11$ has been used in the rest of the experiments.}
    \label{tab:tuningsubgraphsize}
\end{table}

\subsection{Dark Market shutdown: Pre and Post-analysis}
\label{sec:darkmarket}

\textcolor{black}{Dark markets are online platforms operating on the dark web that enable anonymous trade in various goods and services, typically using cryptocurrencies as the medium of exchange. These platforms function outside conventional regulatory systems, allowing users to transact without revealing their identities. When such markets are taken down by law enforcement, it disrupts established transaction patterns and can influence the overall behaviour of cryptocurrency networks \citep{commerce2024darkweb}.} The dark market shutdown, which occurred in the dataset between the 42nd and 43rd day, is one of the events in cryptocurrency markets that causes a change in the nature of transactions \citep{elliptic}. In order to statistically prove this hypothesis, we conducted an analysis, where we studied the top-10 most significant features out of the total 166 features contributing to the class labels before and after the shutdown. We used the chi-square test \citep{chi2} to find the top-10 most significant features for nodes before the dark market shutdown and after the dark market shutdown. Out of 10 features, we found that five features were common before and after the shutdown, while five others were different, showing that 50\% of the top 10 most significant features differ after the dark market shutdown. Hence, to study the distributions of the 5 common features, we focused our statistical analysis on the common features and conducted the Kolmogorov-Smirnov test \citep{kstest} to study the distributions of these 5 common features before and after the dark market shutdown.

Table \ref{tab:stats_summary} describes the mean, standard deviation, skewness and kurtosis of the five common features before and after the dark market shutdown for each category i.e. illicit and licit transactors. The `features' column corresponds to the feature column number. Table \ref{tab:ks_test} describes the p-values corresponding to every class label, as seen, the p-values are extremely small hence we reject the hypothesis that the distributions are the same. Hence, while the 5 most significant features are different after the shutdown, the 5 common features have different distributions after the shutdown, which proves the hypothesis that the nature of transactors changes after the shutdown.

\begin{table*}[htbp]
    \centering
    \renewcommand{\arraystretch}{1.2} 
    \setlength{\tabcolsep}{6pt} 
    \begin{tabular}{llcccc}
        \toprule
        \multirow{2}{*}{Category} & \multirow{2}{*}{Feature} & \multicolumn{4}{c}{Statistics} \\
        \cmidrule(lr){3-6}
        &  & Mean & Stdev & Skewness & Kurtosis \\
        \midrule
        \multirow{5}{*}{\textbf{Illicit Before shutdown}}  
        & 90  & -0.4979  & 0.5385  & 2.4945  & 7.3365  \\
        & 92  & -0.3696  & 0.7572  & 2.9450  & 10.8404  \\
        & 91  & -0.5320  & 0.4295  & 3.5714  & 17.9251  \\
        & 53  & -0.2264  & 0.5459  & 2.7220  & 5.8725  \\
        & 84  & -0.1574  & 0.3211  & 2.7522  & 5.5773  \\
        \midrule
        \multirow{5}{*}{\textbf{Illicit After shutdown}}  
        & 90  & -0.6311  & 0.3138  & 3.3503  & 12.4487  \\
        & 92  & -0.6038  & 0.2797  & 3.3540  & 12.4787  \\
        & 91  & -0.5383  & 0.3799  & 3.3535  & 12.4447  \\
        & 53  & -0.1365  & 0.4059  & 1.4651  & 0.8146  \\
        & 84  & -0.2623  & 0.0005  & 4.7692  & 28.9593  \\
        \midrule
        \multirow{5}{*}{\textbf{Licit Before shutdown}}  
        & 90  & 0.5322  & 1.3463  & 0.6321  & -1.1308  \\
        & 92  & 0.4818  & 1.3502  & 1.0373  & 0.2732  \\
        & 91  & 0.5340  & 1.4393  & 0.9650  & -0.5017  \\
        & 53  & 0.6413  & 1.5406  & 1.4967  & 1.0306  \\
        & 84  & 0.4788  & 1.5634  & 3.1451  & 11.2372  \\
        \midrule
        \multirow{5}{*}{\textbf{Licit After shutdown}}  
        & 90  & 0.2540  & 1.1329  & 1.0261  & -0.0985  \\
        & 92  & 0.1134  & 1.0392  & 1.6857  & 2.6291  \\
        & 91  & 0.1872  & 1.0524  & 1.4194  & 1.4809  \\
        & 53  & 0.5500  & 1.3518  & 1.2808  & 0.4590  \\
        & 84  & 0.3219  & 1.4823  & 3.4683  & 13.2170  \\
        \bottomrule
    \end{tabular}
    \caption{Statistical Summary of Features for Illicit and Licit Categories Before and After the dark market shutdown}
    \label{tab:stats_summary}
\end{table*}

\begin{table*}[htbp]
    \centering
    \renewcommand{\arraystretch}{1.2} 
    \setlength{\tabcolsep}{10pt} 
    \begin{tabular}{lcc}
        \toprule
        \textbf{Feature} & \textbf{p-value (Licit Before vs. After)} & \textbf{p-value (illicit Before vs. After)} \\
        \midrule
        90  & \( 1.6980 \times 10^{-68} \) & \( 9.6456 \times 10^{-13} \) \\
        92  & \( 5.6687 \times 10^{-82} \) & \( 4.3548 \times 10^{-11} \) \\
        91  & \( 2.8609 \times 10^{-61} \) & \( 4.9310 \times 10^{-52} \) \\
        53  & \( 1.5793 \times 10^{-30} \) & \( 9.8974 \times 10^{-30} \) \\
        84  & \( 2.9356 \times 10^{-67} \) & \( 3.1946 \times 10^{-7} \) \\
        \bottomrule
    \end{tabular}
    \caption{KS-test p-values for Illicit and Licit Categories}
    \label{tab:ks_test}
\end{table*}

\subsection{Comparison: Dynamic graph classification models}
\label{sec:compdynamicgraph}

\begin{figure*}[htbp!]
  \centering
  \includegraphics[width=0.75\linewidth]{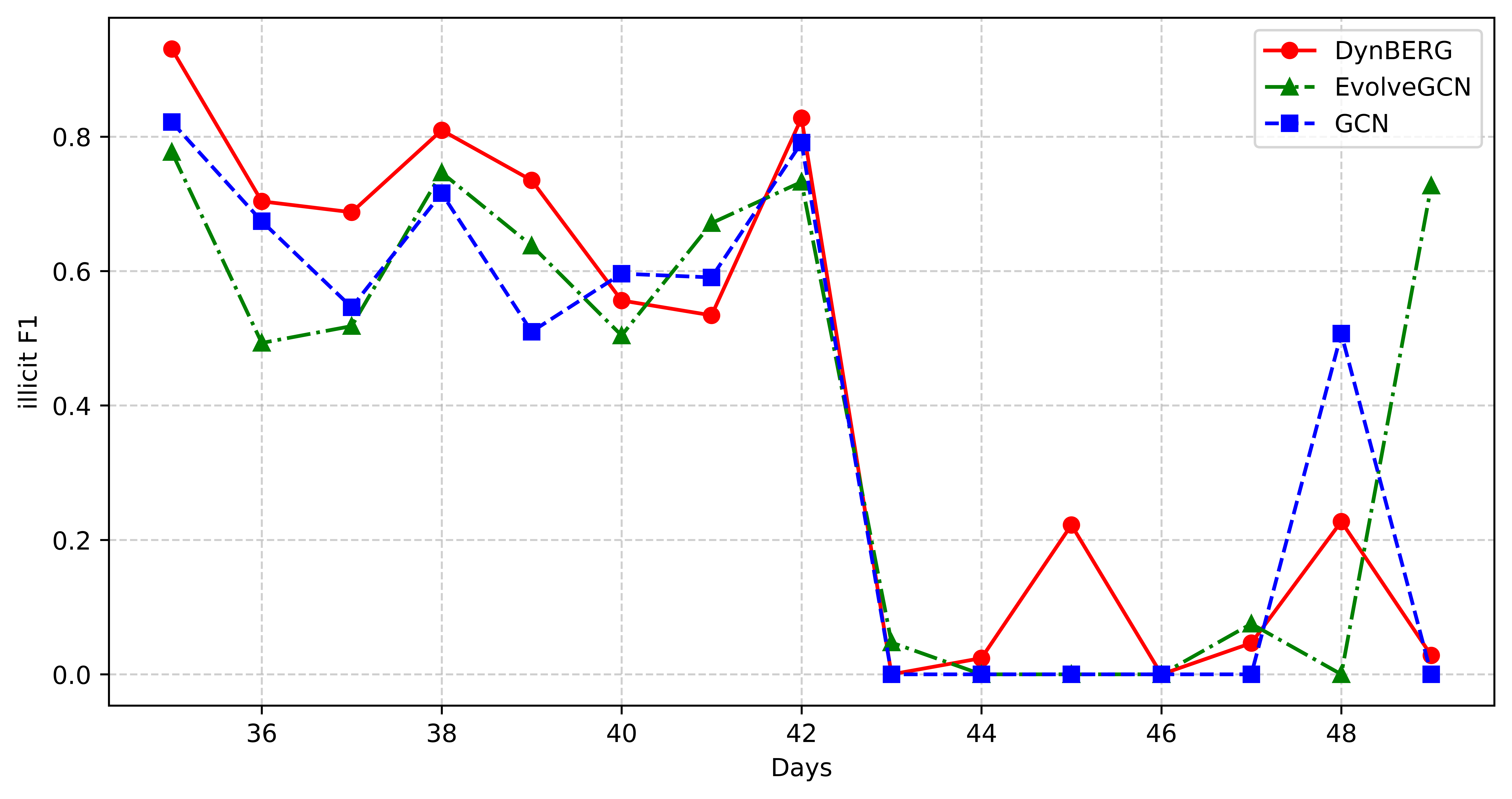}
  \captionof{figure}{Comparison of DynBERG with other state-of-the-art dynamic graph classification models}
  \label{fig:timewisemodelcomp}
\end{figure*}

We next compare the performance of DynBERG with other state-of-the-art dynamic graph classification models like EvolveGCN and GCN. As seen \textcolor{black}{in table \ref{tab:timewisemodelcomp} and figure \ref{fig:timewisemodelcomp}}, the Pre-dark market shutdown performance of DynBERG is much better than that of EvolveGCN and GCN, this can also be confirmed by the high performance of DynBERG for the windows from 35-37 days and 38-40 days. However, after the dark market shutdown, while EvolveGCN recovers by the 48th time step and GCN gives a peak performance at the 49th time step, DynBERG although recovering faster than the other two models at the 45th time step, still faces difficulty in improving performance beyond the 45th time step. The failure of all 3 models to perform better after dark market shutdown can be attributed to the change in the nature of transactions after the shutdown (statistical evidence for which given in \textcolor{black}{table \ref{tab:stats_summary}}) due to which the transactions that were used to train these models changed their nature post the shutdown hence their performance dropped \citep{elliptic}.

\begin{table*}[htbp]
    \centering
    \begin{tabular}{c c c c c}
        \toprule
        & \textbf{DynBERG} & \textbf{EvolveGCN} & \textbf{GCN} & \textbf{Rank} \\
        \midrule
        \textbf{Pre-dark market shutdown} & 0.723(0.1258) & 0.6351(0.1086) & 0.6556(0.1067) & 1\\
        \textbf{Post-dark market shutdown} & 0.0914(0.0953) & 0.1337(0.2668) & 0.0845(0.1889) & 2 \\
        \textbf{35-37} & 0.7739(0.1109) & 0.5961(0.1284) & 0.6806(0.1127) & 1 \\
        \textbf{38-40} & 0.7002(0.1063) & 0.6295(0.0991) & 0.6072(0.0845) & 1 \\
        \textbf{41-43} & 0.4539(0.3426) & 0.4838(0.3096) & 0.4605(0.3358) & 3 \\
        \textbf{44-46} & 0.0821(0.0996) & 0.0(0.0) & 0.0(0.0) & 1 \\
        \textbf{47-49} & 0.1007(0.0898) & 0.2674(0.3264) & 0.169(0.239) & 3 \\
        \bottomrule
    \end{tabular}
    \caption{Pre and Post-dark market shutdown comparison of illicit-F1 performance of DynBERG, EvolveGCN, and GCN}
    \label{tab:timewisemodelcomp}
\end{table*}

\subsection{Ablation Study: Importance of GRU}
\label{sec:ablationstudy}

\begin{figure*}[htbp!]
  \centering
  \includegraphics[width=0.75\linewidth]{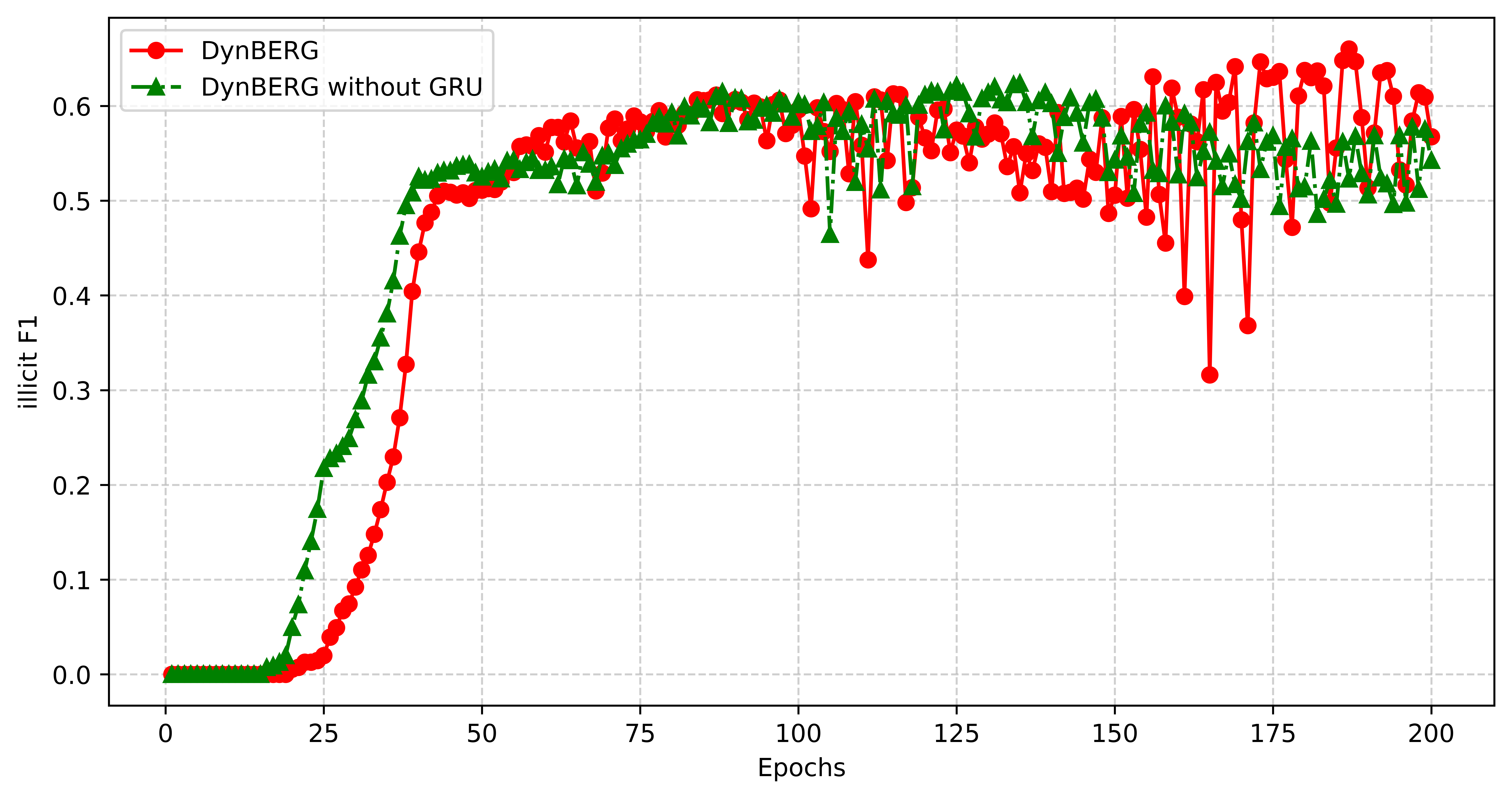}
  \captionof{figure}{illicit F1 vs Epoch comparison of DynBERG vs DynBERG without GRU, highlighting the performance of DynBERG.}
  \label{fig:f1vsepochgrucomp}
\end{figure*}

\begin{figure*}[htbp!]
  \centering
  \includegraphics[width=0.75\linewidth]{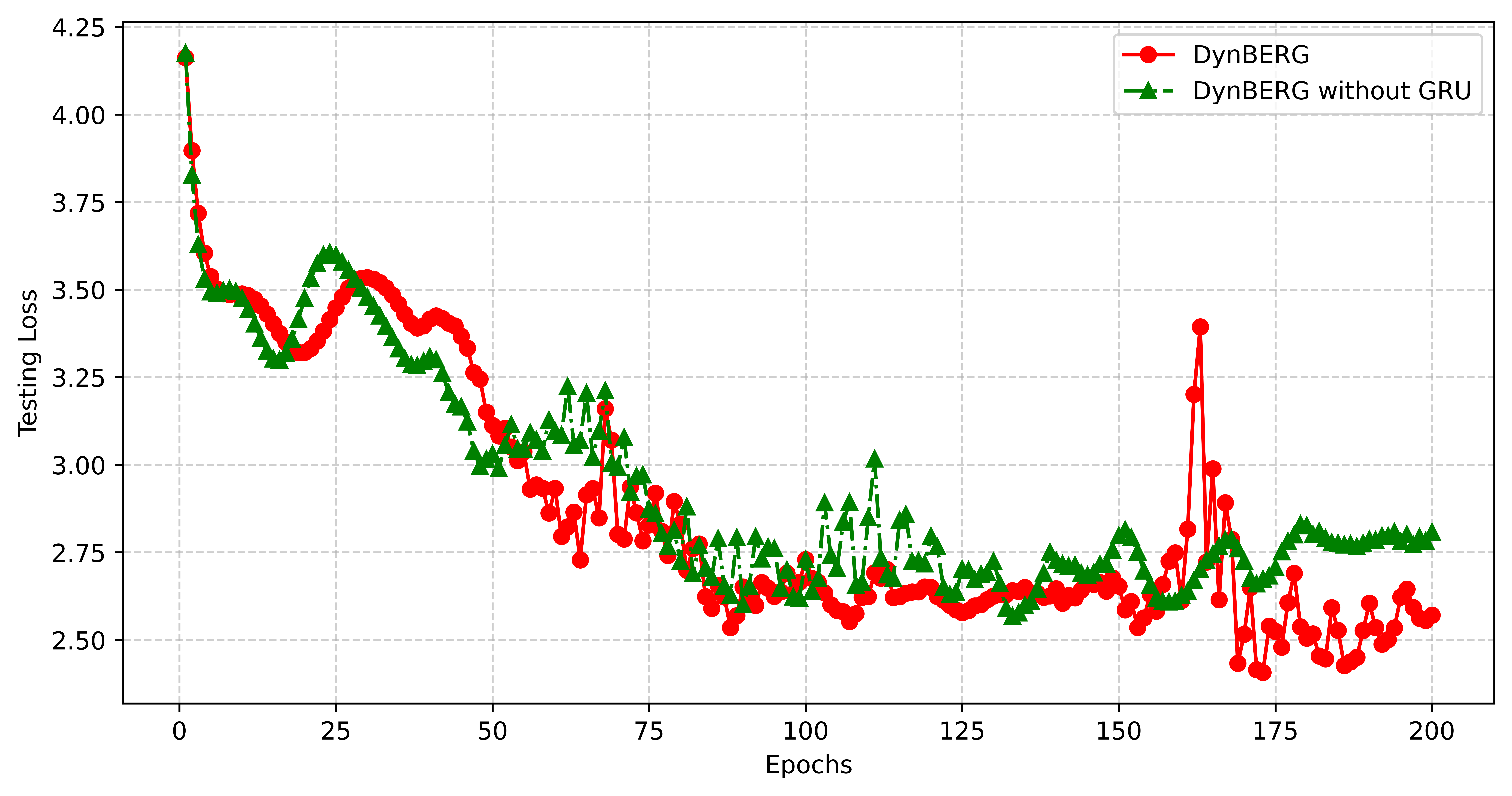}
  \captionof{figure}{Testing loss vs Epoch comparison of DynBERG vs DynBERG without GRU, highlighting the performance of DynBERG.}
  \label{fig:lossvsepochgrucomp}
\end{figure*}

\begin{figure*}[htbp!]
  \centering
  \includegraphics[width=0.75\linewidth]{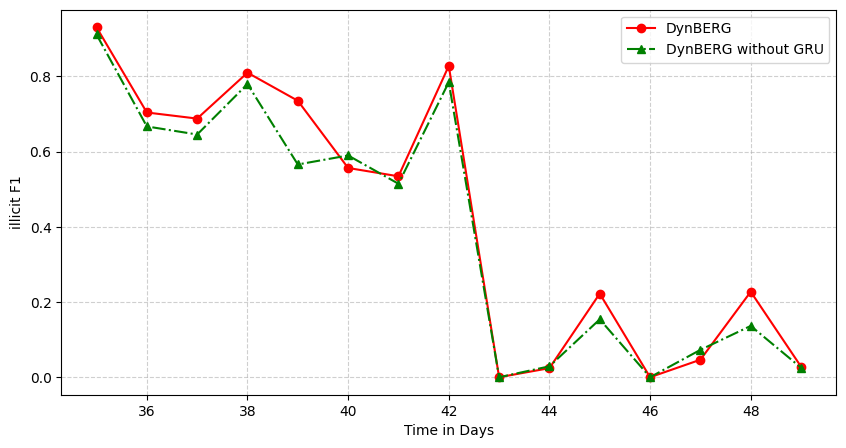}
  \captionof{figure}{Time-wise performance comparison of DynBERG with the same model without GRU layer.}
  \label{fig:timewisegrucomp}
\end{figure*}

We conduct an ablation study to highlight the importance of the time series component of our model. We compare the epoch-wise and testing days-wise performance of DynBERG with thecounterpart, where we remove the GRU layer; therefore, essentially $w_{GRU} = 0 $ and $w_{BERG} = 1$. As seen in Figures \ref{fig:f1vsepochgrucomp} and \ref{fig:lossvsepochgrucomp}, DynBERG trains slower than its counterpart, but it reaches its peak performance that surpasses the overall performance of its counterpart. Beyond 125 epochs, DynBERG without GRUs' performance starts to decay, while DynBERG's performance keeps improving, although with an initial drop. Table \ref{tab:darkmarketgrucomp} further provides information on the performance of the shutdown of the pre- and post-dark markets of both models, as measured by the illicit F1 scores. The analysis made earlier is confirmed by the numerical values of the illicit F1 scores. In the time-wise performance comparison of both models, as shown in figure \ref{fig:timewisegrucomp}, we can see that DynBERG performs better for all 15 testing days leaving out the 40th and 47th day. 

\begin{table*}[htbp]
    \centering
    \resizebox{\textwidth}{!}{
    \begin{tabular}{c c c c c}
        \toprule
        \textbf{Epochs} & \multicolumn{2}{c}{\textbf{DynBERG (Illicit F1 mean(stddev))}} & \multicolumn{2}{c}{\textbf{DynBERG without GRU (Illicit F1 mean(stddev))}} \\
        \cmidrule(lr){2-3} \cmidrule(lr){4-5}
        & Pre Dark Market Shutdown & Post Dark Market Shutdown & Pre Dark Market Shutdown & Post Dark Market Shutdown \\
        \midrule
        20  & 0.0196(0.0457) & 0.0(0.0) & 0.1093(0.0779) & 0.009(0.0201) \\
        40  & 0.5141(0.1227) & 0.0167(0.0373) & 0.5473(0.1255) & 0.0163(0.0364) \\
        60  & 0.632(0.1282) & 0.0325(0.0669) & 0.5925(0.146) & 0.0731(0.0891) \\
        80  & 0.5831(0.2103) & 0.1405(0.1723) & 0.6239(0.1627) & 0.1115(0.1721) \\
        100 & 0.5143(0.2644) & 0.1163(0.1432) & 0.6373(0.1494) & 0.0551(0.0453) \\
        120 & 0.5206(0.2664) & 0.0678(0.1046) & 0.6902(0.1234) & 0.0638(0.0468) \\
        140 & 0.6019(0.2027) & 0.0592(0.0915) & 0.5871(0.2062) & 0.0728(0.0606) \\
        160 & 0.3908(0.2576) & 0.0448(0.0483) & 0.6381(0.1669) & 0.0773(0.0666) \\
        180 & 0.7067(0.1165) & 0.0799(0.1281) & 0.6069(0.2001) & 0.0863(0.0734) \\
        200 & 0.5674(0.2512) & 0.0421(0.0483) & 0.5599(0.2362) & 0.0552(0.0618) \\
        \bottomrule
    \end{tabular}}
    \caption{Comparison of DynBERG and DynBERG without GRU on Illicit F1 scores before and after the Dark Market Shutdown. Highlighting the performance and efficiency of GRU layer to capture performance before and after a major event in the timeline.}
    \label{tab:darkmarketgrucomp}
\end{table*}

\section{Discussion}

We presented   DynBERG, a hybrid of Graph-BERT with a GRU layer which facilitates dynamic graph node classification for financial transactions using the `Elliptic' dataset for money laundering/financial fraud detection. The results highlight the strengths and limitations of DynBERG in dynamic graph classification, particularly in detecting illicit transactions. One of the key findings is the importance of subgraph batch size in optimising model performance where  a batch size of 11 was the most effective, providing the best illicit F1 and micro F1 scores. A smaller batch size may fail to provide enough contextual information, while a larger one could introduce noise and computational inefficiencies. Our statistical analysis of transaction features before and after the dark market shutdown confirms that the nature of transactors changes significantly following such events, as both the most significant features and their distributions exhibit notable shifts.  Our results show that DynBERG outperforms existing state-of-the-art dynamic graph classification models, EvolveGCN and GCN, in the pre-dark market shutdown period, particularly in the evaluation windows spanning 35-37 days and 38-40 days. 

This highlights its effectiveness in modelling illicit transaction patterns under stable conditions. However, following the dark market shutdown, all three models experienced performance degradation (Figure \textcolor{black}{\ref{fig:timewisemodelcomp}}). Although DynBERG recovers faster than its counterparts, it struggles to further enhance its performance beyond this recovery point. In contrast, EvolveGCN and GCN demonstrate slight recoveries at later timesteps (48 and 49) \textcolor{black}{as seen in figure \ref{fig:timewisemodelcomp}}. This suggests that while DynBERG efficiently captures pre-existing illicit behaviour patterns, it may need additional mechanisms to adapt to sudden shifts in transaction dynamics. The observed decline in performance post-shutdown could be attributed to the fundamental change in transaction behaviours, which differs significantly from the data used during training.

The ablation study (Table \ref{tab:darkmarketgrucomp}) provides further insights into the significance of the GRU layer in modelling temporal dependencies. Our results  revealed that removing the GRU layer results in faster initial training; in contrast, DynBERG continues to improve, demonstrating the importance of capturing long-term dependencies in illicit transaction patterns. Moreover, in the testing day-wise analysis, DynBERG consistently outperforms its counterpart across most testing days. These findings confirm that the GRU layer plays a crucial role in ensuring model stability and robustness over extended periods.


Despite its advantages, DynBERG faces challenges in adapting to rapidly evolving transaction patterns, particularly in response to external disruptions such as regulatory actions or market shutdowns. The observed performance drop post-shutdown suggests that the model relies heavily on historical transaction structures, which may limit its adaptability in scenarios where financial behaviors shift unpredictably. \textcolor{black}{Furthermore, the use of the average pooling layer might cause the loss of important information encoded in the final states output by the transformer which also might be the reason for limited adaptibility of the model}. To address this limitation, future work could explore adaptive learning strategies, such as self-supervised pre-training, reinforcement learning, or domain adaptation techniques, to help the model adjust to changing data distributions. 

These findings have important implications for real-world financial crime detection. The strong performance of DynBERG in stable conditions suggests its potential as a valuable tool for detecting illicit transactions in pre-identified financial networks. However, its post-shutdown performance challenges underscore the need for continuous model monitoring and adaptation. In practical applications, deploying models that can dynamically retrain using recent transactional data may improve long-term effectiveness. Future research should focus on developing systems that can detect and respond to changes in fraudulent transaction patterns in real-time, ensuring their continued relevance in an ever-evolving financial landscape.

\section{Conclusion}
In this work, we introduced DynBERG, a hybrid model that combines Graph-BERT with a GRU layer to handle node classification on dynamic graphs. We tested it on the Elliptic dataset, which is used for detecting financial fraud, and found that DynBERG performs well in stable conditions. It was able to capture transaction patterns better than existing models like EvolveGCN and GCN before major disruptions. Our experiments also showed that the size of the subgraph batches and the GRU layer are both important for good performance, as they help the model capture context and time-related information.

Our results show that DynBERG outperforms EvolveGCN before the Dark Market Shutdown and beats GCN after the event. Additionally, we conduct an ablation study to highlight the importance of incorporating a time-series deep learning layer like GRU. This ablation study compares DynBERG with a variant that omits the GRU layer, demonstrating the crucial role of temporal modelling in financial transaction analysis.

However, DynBERG struggles when the transaction patterns change suddenly, such as after a dark market shutdown. This shows that the model depends a lot on past behavior and has trouble adjusting when the data distribution changes quickly. To improve this, future work could explore more adaptive learning approaches like self-supervised learning or methods that let the model update itself in real-time.

Overall, DynBERG shows strong potential for fraud detection in financial networks, especially when the environment is stable. But to make it useful in real-world scenarios, it will need better ways to adapt to changing behaviours over time.




\section*{Code and Data Availability}

GitHub repository \url{}.

\bibliography{refs}   

\begin{thebibliography}{10}
\expandafter\ifx\csname url\endcsname\relax
  \def\url#1{\texttt{#1}}\fi
\expandafter\ifx\csname urlprefix\endcsname\relax\def\urlprefix{URL }\fi
\expandafter\ifx\csname href\endcsname\relax
  \def\href#1#2{#2} \def\path#1{#1}\fi

\bibitem{Yu2018GraphTheoryBrain}
Q.~Yu, Y.~Du, J.~Chen, J.~Sui, T.~Adali, G.~Pearlson, V.~D. Calhoun, \href{https://doi.org/10.1109/JPROC.2018.2825200}{Application of graph theory to assess static and dynamic brain connectivity: Approaches for building brain graphs}, Proceedings of the IEEE 106~(5) (2018) 886--906, epub 2018 Apr 25.
\newblock \href {https://doi.org/10.1109/JPROC.2018.2825200} {\path{doi:10.1109/JPROC.2018.2825200}}.
\newline\urlprefix\url{https://doi.org/10.1109/JPROC.2018.2825200}

\bibitem{ugander2011anatomyfacebooksocialgraph}
J.~Ugander, B.~Karrer, L.~Backstrom, C.~Marlow, \href{https://arxiv.org/abs/1111.4503}{The anatomy of the facebook social graph} (2011).
\newblock \href {http://arxiv.org/abs/1111.4503} {\path{arXiv:1111.4503}}.
\newline\urlprefix\url{https://arxiv.org/abs/1111.4503}

\bibitem{Huber2007GraphsMolecularBiology}
W.~Huber, V.~J. Carey, L.~Long, S.~Falcon, R.~Gentleman, \href{https://doi.org/10.1186/1471-2105-8-S6-S8}{Graphs in molecular biology}, BMC Bioinformatics 8~(6) (2007) S8.
\newblock \href {https://doi.org/10.1186/1471-2105-8-S6-S8} {\path{doi:10.1186/1471-2105-8-S6-S8}}.
\newline\urlprefix\url{https://doi.org/10.1186/1471-2105-8-S6-S8}

\bibitem{laclau2024surveyfairnessmachinelearning}
C.~Laclau, C.~Largeron, M.~Choudhary, \href{https://arxiv.org/abs/2205.05396}{A survey on fairness for machine learning on graphs} (2024).
\newblock \href {http://arxiv.org/abs/2205.05396} {\path{arXiv:2205.05396}}.
\newline\urlprefix\url{https://arxiv.org/abs/2205.05396}

\bibitem{GraRep}
S.~Cao, W.~Lu, Q.~Xu, \href{https://doi.org/10.1145/2806416.2806512}{Grarep: Learning graph representations with global structural information}, in: Proceedings of the 24th ACM International on Conference on Information and Knowledge Management, CIKM '15, Association for Computing Machinery, New York, NY, USA, 2015, p. 891–900.
\newblock \href {https://doi.org/10.1145/2806416.2806512} {\path{doi:10.1145/2806416.2806512}}.
\newline\urlprefix\url{https://doi.org/10.1145/2806416.2806512}

\bibitem{kipfwellingGCN}
T.~N. Kipf, M.~Welling, \href{http://arxiv.org/abs/1609.02907}{Semi-supervised classification with graph convolutional networks}, CoRR abs/1609.02907 (2016).
\newblock \href {http://arxiv.org/abs/1609.02907} {\path{arXiv:1609.02907}}.
\newline\urlprefix\url{http://arxiv.org/abs/1609.02907}

\bibitem{graphattentionnetworks}
P.~Velickovic, G.~Cucurull, A.~Casanova, A.~Romero, P.~Liò, Y.~Bengio, \href{https://arxiv.org/abs/1710.10903}{Graph attention networks} (2018).
\newblock \href {http://arxiv.org/abs/1710.10903} {\path{arXiv:1710.10903}}.
\newline\urlprefix\url{https://arxiv.org/abs/1710.10903}

\bibitem{graphunets}
H.~Gao, S.~Ji, \href{http://arxiv.org/abs/1905.05178}{Graph u-nets}, CoRR abs/1905.05178 (2019).
\newblock \href {http://arxiv.org/abs/1905.05178} {\path{arXiv:1905.05178}}.
\newline\urlprefix\url{http://arxiv.org/abs/1905.05178}

\bibitem{dlfrauddetect}
Y.~Chen, C.~Zhao, Y.~Xu, C.~Nie, Y.~Zhang, \href{https://www.sciencedirect.com/science/article/pii/S2666764925000372}{Deep learning in financial fraud detection: Innovations, challenges, and applications}, Data Science and Management (2025).
\newblock \href {https://doi.org/https://doi.org/10.1016/j.dsm.2025.08.002} {\path{doi:https://doi.org/10.1016/j.dsm.2025.08.002}}.
\newline\urlprefix\url{https://www.sciencedirect.com/science/article/pii/S2666764925000372}

\bibitem{dlfrauddetect2}
L.~Hernandez~Aros, L.~X. Bustamante~Molano, F.~Gutierrez-Portela, J.~J. Moreno~Hernandez, M.~S. Rodr{\'i}guez~Barrero, \href{https://doi.org/10.1057/s41599-024-03606-0}{Financial fraud detection through the application of machine learning techniques: a literature review}, Humanities and Social Sciences Communications 11~(1) (2024) 1130.
\newblock \href {https://doi.org/10.1057/s41599-024-03606-0} {\path{doi:10.1057/s41599-024-03606-0}}.
\newline\urlprefix\url{https://doi.org/10.1057/s41599-024-03606-0}

\bibitem{bitcoin}
C.~Rahalkar, A.~Virgaonkar, \href{https://arxiv.org/abs/2109.07634}{Summarizing and analyzing the privacy-preserving techniques in bitcoin and other cryptocurrencies}, CoRR abs/2109.07634, withdrawn. (2021).
\newblock \href {http://arxiv.org/abs/2109.07634} {\path{arXiv:2109.07634}}.
\newline\urlprefix\url{https://arxiv.org/abs/2109.07634}

\bibitem{graphanomalystatic}
H.~Han, R.~Wang, Y.~Chen, K.~Xie, K.~Zhang, Research on abnormal transaction detection method for blockchain, in: D.~Svetinovic, Y.~Zhang, X.~Luo, X.~Huang, X.~Chen (Eds.), Blockchain and Trustworthy Systems, Springer Nature Singapore, Singapore, 2022, pp. 223--236.

\bibitem{graphanomalystatic2}
V.~Patel, L.~Pan, S.~Rajasegarar, Graph deep learning based anomaly detection in ethereum blockchain network, in: M.~Kuty{\l}owski, J.~Zhang, C.~Chen (Eds.), Network and System Security, Springer International Publishing, Cham, 2020, pp. 132--148.

\bibitem{graphanomalystatic3}
R.~Tan, Q.~Tan, P.~Zhang, Z.~Li, Graph neural network for ethereum fraud detection, in: 2021 IEEE International Conference on Big Knowledge (ICBK), 2021, pp. 78--85.
\newblock \href {https://doi.org/10.1109/ICKG52313.2021.00020} {\path{doi:10.1109/ICKG52313.2021.00020}}.

\bibitem{graphanomalystatic4}
A.~Li, Z.~Wang, M.~Yu, D.~Chen, Blockchain abnormal transaction detection method based on weighted sampling neighborhood nodes, in: 2022 3rd International Conference on Big Data, Artificial Intelligence and Internet of Things Engineering (ICBAIE), 2022, pp. 746--752.
\newblock \href {https://doi.org/10.1109/ICBAIE56435.2022.9985815} {\path{doi:10.1109/ICBAIE56435.2022.9985815}}.

\bibitem{graphanomalydynamic}
V.~Patel, S.~Rajasegarar, L.~Pan, J.~Liu, L.~Zhu, Evangcn: Evolving graph deep neural network based anomaly detection in blockchain, in: W.~Chen, L.~Yao, T.~Cai, S.~Pan, T.~Shen, X.~Li (Eds.), Advanced Data Mining and Applications, Springer Nature Switzerland, Cham, 2022, pp. 444--456.

\bibitem{graphanomalydynamic2}
X.~Wang, D.~Lyu, M.~Li, Y.~Xia, Q.~Yang, X.~Wang, X.~Wang, P.~Cui, Y.~Yang, B.~Sun, Z.~Guo, \href{https://doi.org/10.1145/3448016.3457564}{Apan: Asynchronous propagation attention network for real-time temporal graph embedding}, in: Proceedings of the 2021 International Conference on Management of Data, SIGMOD '21, Association for Computing Machinery, New York, NY, USA, 2021, p. 2628–2638.
\newblock \href {https://doi.org/10.1145/3448016.3457564} {\path{doi:10.1145/3448016.3457564}}.
\newline\urlprefix\url{https://doi.org/10.1145/3448016.3457564}

\bibitem{evolvegcn}
A.~Pareja, G.~Domeniconi, J.~Chen, T.~Ma, T.~Suzumura, H.~Kanezashi, T.~Kaler, C.~E. Leiserson, \href{http://arxiv.org/abs/1902.10191}{Evolvegcn: Evolving graph convolutional networks for dynamic graphs}, CoRR abs/1902.10191 (2019).
\newblock \href {http://arxiv.org/abs/1902.10191} {\path{arXiv:1902.10191}}.
\newline\urlprefix\url{http://arxiv.org/abs/1902.10191}

\bibitem{oversmoothing}
Q.~Li, Z.~Han, X.~Wu, \href{http://arxiv.org/abs/1801.07606}{Deeper insights into graph convolutional networks for semi-supervised learning}, CoRR abs/1801.07606 (2018).
\newblock \href {http://arxiv.org/abs/1801.07606} {\path{arXiv:1801.07606}}.
\newline\urlprefix\url{http://arxiv.org/abs/1801.07606}

\bibitem{gco}
D.~K. Hammond, P.~Vandergheynst, R.~Gribonval, \href{https://www.sciencedirect.com/science/article/pii/S1063520310000552}{Wavelets on graphs via spectral graph theory}, Applied and Computational Harmonic Analysis 30~(2) (2011) 129--150.
\newblock \href {https://doi.org/https://doi.org/10.1016/j.acha.2010.04.005} {\path{doi:https://doi.org/10.1016/j.acha.2010.04.005}}.
\newline\urlprefix\url{https://www.sciencedirect.com/science/article/pii/S1063520310000552}

\bibitem{graphbert}
J.~Zhang, H.~Zhang, C.~Xia, L.~Sun, \href{https://arxiv.org/abs/2001.05140}{Graph-bert: Only attention is needed for learning graph representations}, CoRR abs/2001.05140 (2020).
\newblock \href {http://arxiv.org/abs/2001.05140} {\path{arXiv:2001.05140}}.
\newline\urlprefix\url{https://arxiv.org/abs/2001.05140}

\bibitem{bert1}
J.~Devlin, M.~Chang, K.~Lee, K.~Toutanova, \href{http://arxiv.org/abs/1810.04805}{{BERT:} pre-training of deep bidirectional transformers for language understanding}, CoRR abs/1810.04805 (2018).
\newblock \href {http://arxiv.org/abs/1810.04805} {\path{arXiv:1810.04805}}.
\newline\urlprefix\url{http://arxiv.org/abs/1810.04805}

\bibitem{gru}
K.~Cho, B.~van Merri{\"e}nboer, D.~Bahdanau, Y.~Bengio, \href{https://aclanthology.org/W14-4012/}{On the properties of neural machine translation: Encoder{--}decoder approaches}, in: D.~Wu, M.~Carpuat, X.~Carreras, E.~M. Vecchi (Eds.), Proceedings of {SSST}-8, Eighth Workshop on Syntax, Semantics and Structure in Statistical Translation, Association for Computational Linguistics, Doha, Qatar, 2014, pp. 103--111.
\newblock \href {https://doi.org/10.3115/v1/W14-4012} {\path{doi:10.3115/v1/W14-4012}}.
\newline\urlprefix\url{https://aclanthology.org/W14-4012/}

\bibitem{elliptic}
M.~Weber, G.~Domeniconi, J.~Chen, D.~K.~I. Weidele, C.~Bellei, T.~Robinson, C.~E. Leiserson, \href{http://arxiv.org/abs/1908.02591}{Anti-money laundering in bitcoin: Experimenting with graph convolutional networks for financial forensics}, CoRR abs/1908.02591 (2019).
\newblock \href {http://arxiv.org/abs/1908.02591} {\path{arXiv:1908.02591}}.
\newline\urlprefix\url{http://arxiv.org/abs/1908.02591}

\bibitem{matrixfact1}
M.~Belkin, P.~Niyogi, \href{https://proceedings.neurips.cc/paper_files/paper/2001/file/f106b7f99d2cb30c3db1c3cc0fde9ccb-Paper.pdf}{Laplacian eigenmaps and spectral techniques for embedding and clustering}, in: T.~Dietterich, S.~Becker, Z.~Ghahramani (Eds.), Advances in Neural Information Processing Systems, Vol.~14, MIT Press, 2001, pp. 585--591.
\newline\urlprefix\url{https://proceedings.neurips.cc/paper_files/paper/2001/file/f106b7f99d2cb30c3db1c3cc0fde9ccb-Paper.pdf}

\bibitem{matrixfact2}
J.~Li, H.~Dani, X.~Hu, J.~Tang, Y.~Chang, H.~Liu, \href{http://arxiv.org/abs/1706.01860}{Attributed network embedding for learning in a dynamic environment}, CoRR abs/1706.01860 (2017).
\newblock \href {http://arxiv.org/abs/1706.01860} {\path{arXiv:1706.01860}}.
\newline\urlprefix\url{http://arxiv.org/abs/1706.01860}

\bibitem{randomwalk2}
A.~Grover, J.~Leskovec, \href{https://doi.org/10.1145/2939672.2939754}{node2vec: Scalable feature learning for networks}, in: Proceedings of the 22nd ACM SIGKDD International Conference on Knowledge Discovery and Data Mining, KDD '16, Association for Computing Machinery, New York, NY, USA, 2016, p. 855–864.
\newblock \href {https://doi.org/10.1145/2939672.2939754} {\path{doi:10.1145/2939672.2939754}}.
\newline\urlprefix\url{https://doi.org/10.1145/2939672.2939754}

\bibitem{randomwalk1}
G.~H. Nguyen, J.~B. Lee, R.~A. Rossi, N.~K. Ahmed, E.~Koh, S.~Kim, \href{https://doi.org/10.1145/3184558.3191526}{Continuous-time dynamic network embeddings}, in: Companion Proceedings of the The Web Conference 2018, WWW '18, International World Wide Web Conferences Steering Committee, Republic and Canton of Geneva, CHE, 2018, p. 969–976.
\newblock \href {https://doi.org/10.1145/3184558.3191526} {\path{doi:10.1145/3184558.3191526}}.
\newline\urlprefix\url{https://doi.org/10.1145/3184558.3191526}

\bibitem{deeplearning1}
P.~Goyal, N.~Kamra, X.~He, Y.~Liu, \href{http://arxiv.org/abs/1805.11273}{Dyngem: Deep embedding method for dynamic graphs}, CoRR abs/1805.11273 (2018).
\newblock \href {http://arxiv.org/abs/1805.11273} {\path{arXiv:1805.11273}}.
\newline\urlprefix\url{http://arxiv.org/abs/1805.11273}

\bibitem{deeplearning2}
R.~Trivedi, H.~Dai, Y.~Wang, L.~Song, \href{http://arxiv.org/abs/1705.05742}{Know-evolve: Deep reasoning in temporal knowledge graphs}, CoRR abs/1705.05742 (2017).
\newblock \href {http://arxiv.org/abs/1705.05742} {\path{arXiv:1705.05742}}.
\newline\urlprefix\url{http://arxiv.org/abs/1705.05742}

\bibitem{deeplearning3}
R.~Trivedi, M.~Farajtabar, P.~Biswal, H.~Zha, \href{http://arxiv.org/abs/1803.04051}{Representation learning over dynamic graphs}, CoRR abs/1803.04051 (2018).
\newblock \href {http://arxiv.org/abs/1803.04051} {\path{arXiv:1803.04051}}.
\newline\urlprefix\url{http://arxiv.org/abs/1803.04051}

\bibitem{deeplearning4}
Y.~Zuo, G.~Liu, H.~Lin, J.~Guo, X.~Hu, J.~Wu, \href{https://doi.org/10.1145/3219819.3220054}{Embedding temporal network via neighborhood formation}, in: Proceedings of the 24th ACM SIGKDD International Conference on Knowledge Discovery \& Data Mining, KDD '18, Association for Computing Machinery, New York, NY, USA, 2018, p. 2857–2866.
\newblock \href {https://doi.org/10.1145/3219819.3220054} {\path{doi:10.1145/3219819.3220054}}.
\newline\urlprefix\url{https://doi.org/10.1145/3219819.3220054}

\bibitem{deeplearning5}
Y.~Seo, M.~Defferrard, P.~Vandergheynst, X.~Bresson, \href{https://arxiv.org/abs/1612.07659}{Structured sequence modeling with graph convolutional recurrent networks} (2016).
\newblock \href {http://arxiv.org/abs/1612.07659} {\path{arXiv:1612.07659}}.
\newline\urlprefix\url{https://arxiv.org/abs/1612.07659}

\bibitem{deeplearning6}
F.~Manessi, A.~Rozza, M.~Manzo, \href{http://arxiv.org/abs/1704.06199}{Dynamic graph convolutional networks}, CoRR abs/1704.06199 (2017).
\newblock \href {http://arxiv.org/abs/1704.06199} {\path{arXiv:1704.06199}}.
\newline\urlprefix\url{http://arxiv.org/abs/1704.06199}

\bibitem{reviewpaperfin1}
S.~Motie, B.~Raahemi, \href{https://www.sciencedirect.com/science/article/pii/S0957417423026581}{Financial fraud detection using graph neural networks: A systematic review}, Expert Systems with Applications 240 (2024) 122156.
\newblock \href {https://doi.org/https://doi.org/10.1016/j.eswa.2023.122156} {\path{doi:https://doi.org/10.1016/j.eswa.2023.122156}}.
\newline\urlprefix\url{https://www.sciencedirect.com/science/article/pii/S0957417423026581}

\bibitem{normadjpaper}
L.~Shi, Y.~Zhang, J.~Cheng, H.~Lu, Skeleton-based action recognition with directed graph neural networks, in: 2019 IEEE/CVF Conference on Computer Vision and Pattern Recognition (CVPR), 2019, pp. 7904--7913.
\newblock \href {https://doi.org/10.1109/CVPR.2019.00810} {\path{doi:10.1109/CVPR.2019.00810}}.

\bibitem{gres}
J.~Zhang, L.~Meng, \href{http://arxiv.org/abs/1909.05729}{Gresnet: Graph residual network for reviving deep gnns from suspended animation}, CoRR abs/1909.05729 (2019).
\newblock \href {http://arxiv.org/abs/1909.05729} {\path{arXiv:1909.05729}}.
\newline\urlprefix\url{http://arxiv.org/abs/1909.05729}

\bibitem{commerce2024darkweb}
A.~Nishnianidze, Commerce in the shadows: Exploring dark web black markets, Law and World 10~(30) (2024) 193--216.
\newblock \href {https://doi.org/10.36475/10.2.14} {\path{doi:10.36475/10.2.14}}.

\bibitem{chi2}
M.~L. McHugh, The chi-square test of independence, Biochemia Medica (Zagreb) 23 (2013) 143--149.
\newblock \href {https://doi.org/10.11613/BM.2013.018} {\path{doi:10.11613/BM.2013.018}}.

\bibitem{kstest}
F.~J. Massey, \href{http://www.jstor.org/stable/2280095}{The kolmogorov-smirnov test for goodness of fit}, Journal of the American Statistical Association 46~(253) (1951) 68--78.
\newline\urlprefix\url{http://www.jstor.org/stable/2280095}

\end{thebibliography}
\end{document}